\documentclass[a4paper,conference]{IEEEtran}

\IEEEoverridecommandlockouts 

\usepackage{amsmath}
\usepackage{amssymb} 
\usepackage{bbm}
\usepackage{dsfont}
\usepackage{mathtools}

\usepackage{graphicx}
\graphicspath{ {./images/} }

\title{\LARGE \bf
Individual common dolphin identification via metric embedding learning
}

  \author{
    \IEEEauthorblockN{Soren BOUMA\IEEEauthorrefmark{1}, Matthew D. M. PAWLEY\IEEEauthorrefmark{1}, Krista HUPMAN\IEEEauthorrefmark{2} and Andrew GILMAN\IEEEauthorrefmark{1}}
    \IEEEauthorblockA{\IEEEauthorrefmark{1}Institute of Natural and Mathematical Sciences, Massey University, Auckland, New Zealand \\
    \IEEEauthorrefmark{2}NIWA, Wellington, New Zealand
    \\a.gilman@massey.ac.nz}
  }

\begin{document}

\IEEEoverridecommandlockouts
\IEEEpubid{\makebox[\columnwidth]{978-1-7281-0125-5/18/\$31.00 \textcopyright{}2018 IEEE \hfill}
\hspace{\columnsep}\makebox[\columnwidth]{ }}

\maketitle
\IEEEpubidadjcol

\newcommand{\nnfn}{f_\theta}

\begin{abstract}

Photo-identification (photo-id) of dolphin individuals is a commonly used technique in ecological sciences to monitor state and health of individuals, as well as to study the social structure and distribution of a population. Traditional photo-id involves a laborious manual process of matching each dolphin fin photograph captured in the field to a catalogue of known individuals.

We examine this problem in the context of open-set recognition and utilise a triplet loss function to learn a compact representation of fin images in a Euclidean embedding, where the Euclidean distance metric represents fin similarity. We show that this compact representation can be successfully learnt from a fairly small (in deep learning context) training set and still generalise well to out-of-sample identities (completely new dolphin individuals), with top-1 and top-5 test set (37 individuals) accuracy of $90.5\pm2$ and $93.6\pm1$ percent. In the presence of 1200 distractors, top-1 accuracy dropped by $12\%$; however, top-5 accuracy saw only a $2.8\%$ drop.

\end{abstract}

\section{INTRODUCTION}

Dolphin photo-identification (photo-ID) studies involve photographing dolphin dorsal fins during field work and then having a human categorise images into unique individual animals and matching them with an existing catalogue of known individuals. Individuals are identified by natural features that can be observed on the fins---these features vary between species, but typically include the pattern of nicks and notches on the trailing edge of the fin, the scratches/rake marks/scars on the fin and (for some species) the pigmentation patterns \cite{Wursig1977photographic}. 

Matching of new images from the field against a large catalogue is time consuming, because it requires a human to compare each candidate image to every fin in the catalogue, taking $\mathcal{O}(mn)$ average time (where $m$ is the number of new images and $n$ is the number of catalogued individuals). Moreover, as $n$ increases over time as more individuals are added to the catalogue, so does the likelihood of making a mistake, putting catalogue integrity at risk. Consequently, photo-ID studies don't easily scale to large populations or long periods of time \cite{Pawley2018ExaminingDelphinids, Hupman2018ChallengesDelphinids}. 

The use of machine learning methods to automate the catalogue process (by automatically clustering and/or classifying images) has the potential to make the study of large populations viable; however this is not a trivial task. Training a robust classifier is challenging due to the presence of major variation in appearance (due to variable illumination and pose, out-of-focus blurring, motion blurring, occlusions and specular highlights), discriminative features between individual animals being relatively subtle and a low sample per class rate for many classes in the existing datasets.

Over the past decade, deep learning methods [DL] (with sufficient training data) have consistently outperformed previous state-of-the-art machine learning techniques in the field of computer vision. Convolutional Neural Networks (CNNs) were inspired by the human visual system \cite{LeCun2015DeepLearning} and have been extremely successful in applications such as object recognition \cite{Krizhevsky2012ImageNetNetworks} and detection \cite{Ren2017FasterNetworks}.

Animal photo-id \cite{Kuhl2013AnimalAppearance} is closely related to facial recognition and person re-identification tasks \cite{Zheng2016PersonFuture}, in that it is also an open set recognition problem \cite{Scheirer2013TowardRecognition}. Unlike standard classification, these tasks must generalise to identities outside the training set. DL approaches that have been used to this end include training a multi-class classifier with available identities, whilst using an intermediate bottleneck layer as a generalised representation \cite{Taigman2014DeepFace:Verification,Sun2015DeeplyRobust}, and various metric (embedding) learning approaches \cite{Schroff2015FaceNet:Clustering,Tadmor2016LearningMethod,Hermans2017InRe-Identification,Parkhi2015DeepRecognition}. Of these approaches, a form of metric learning based on triplet loss is of most interest, as it has worked exceptionally well for facial recognition \cite{Schroff2015FaceNet:Clustering} and person re-id tasks \cite{Hermans2017InRe-Identification}, demonstrating state-of-the-art performance on standard datasets. The resulting models showed robustness to variation in pose, lighting, occlusion and camera angle. Another advantage of this approach is that it has given good results despite the training data having strong class imbalances (a common occurrence in some ecological datasets).

\begin{figure}
\centering

\includegraphics[width=\columnwidth]{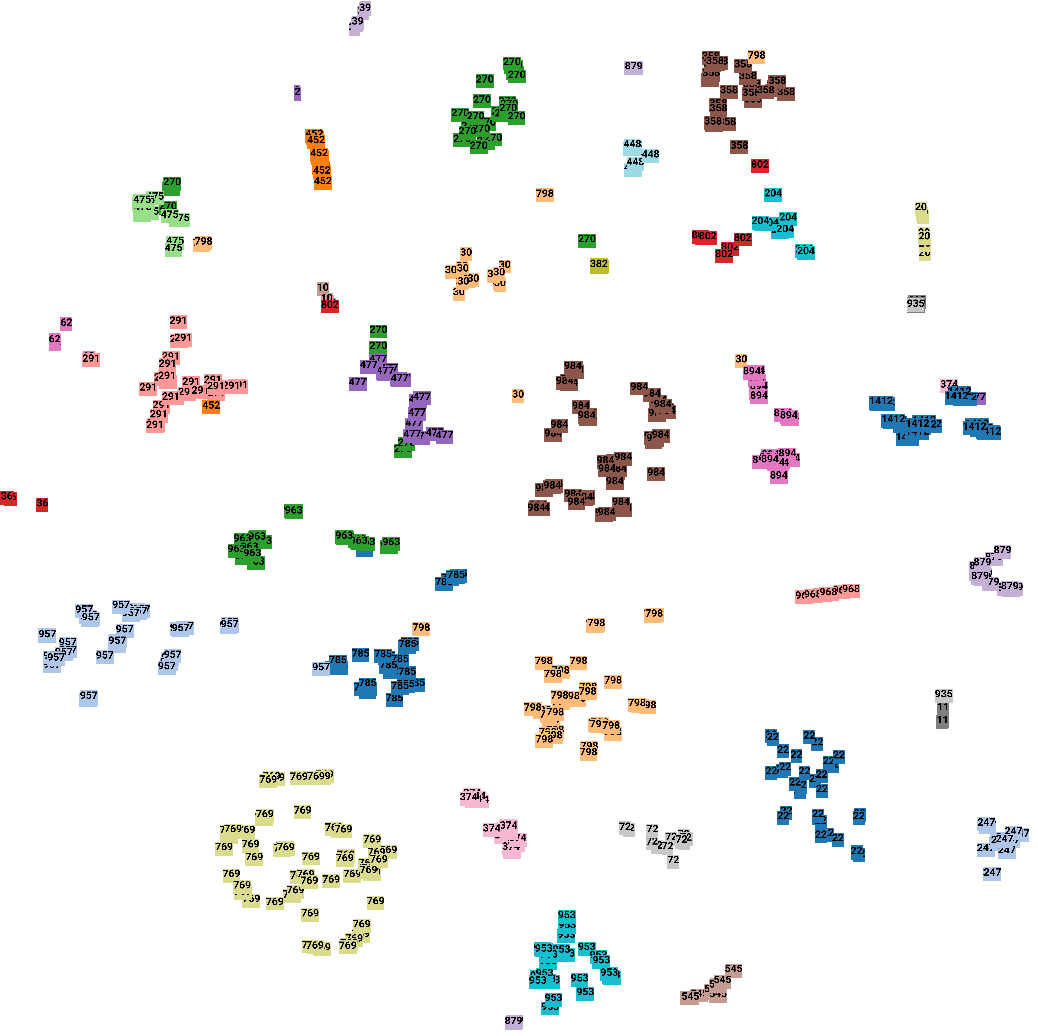}
\caption{Visualisation of the evaluation set embeddings, projected into two dimensions using t-SNE shows good correspondence between clusters and dolphin identities (best viewed zoomed in).}
\label{fig:t-sne}
\end{figure}

The main challenge with applying these methods, and DL methods in general, to animal photo-id is the scarcity of training data. For example, Google's FaceNet \cite{Schroff2015FaceNet:Clustering} was trained on 200M images containing 8M identities. Parkhi et al. \cite{Parkhi2015DeepRecognition} achieved comparable results with 2.6M images containing 2.6K identities; however, the size of that training data is still several orders of magnitude greater than the number of images stored within even large dolphin photo-id catalogues \cite{Pawley2018ExaminingDelphinids}.

To investigate how computer vision and DL can be used to assist ecologists when creating a catalogue in photo-id studies, we have methodically collected dolphin images from available sources and built a medium-sized dataset that could potentially be used to train DL models.

This paper presents our preliminary work trying to answer the following questions:
\begin{itemize}
\item Can a triplet-loss based CNN learn robust discriminative features of dolphin fins that generalise to new identities? %
\item What sort of test accuracy can we expect and what effect does the training set size have on it?
\end{itemize}

\section{METHODS}
\subsection{Dataset}
\begin{figure}[h!]
\centering
\includegraphics[width=0.95\columnwidth,trim={10px 10px 15px 10px},clip]{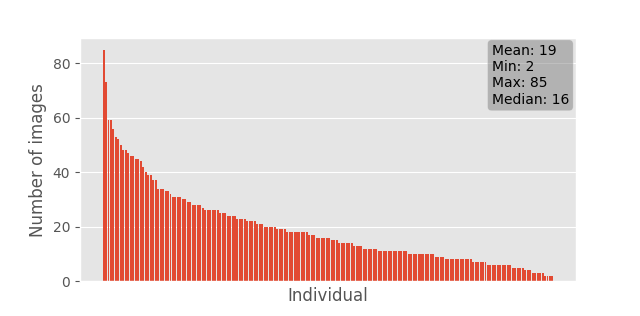}
\caption{Distribution of images per dolphin.}
\label{fig:image_count_hist}
\end{figure}

Image catalogues from photo-id studies are rarely made publicly available and collecting large amounts of training data from different sources can be a challenging task. We were able to collect a dataset containing 3544 images of 185 individual New Zealand common dolphins ($Delphinus$ spp). A further 1,200 singleton images (i.e. of 1,200 different individuals) were also collected. The distribution of images per individual is highly imbalanced in this dataset (see Fig. \ref{fig:image_count_hist}). Moreover, strong correlations are also present between images of the same individual taken during the same encounter (on the same day).

\subsection{Fin Detector}
Consistent with how this method would be used in practice, we cropped fins from images of dolphins using an automated method, available from PhotoId.Ninja \cite{PhotoApplication}: an object detector based on GoogLeNet (v3) \cite{Szegedy2016RethinkingVision} backbone network. This method typically extracts tight crops of fins as shown in Fig. \ref{fig:fin-rankings}.

\subsection{Learning an embedding}
Our approach uses a deep convolutional neural network to learn a Euclidean embedding of cropped dolphin fin images (see Fig. \ref{fig:t-sne} for an example of such an embedding). With each layer, the network transforms the image data, creating a new, more abstract, representation of the data and finally mapping it into a more compact embedding space, such that the Euclidean distances in this space corresponds to fin similarity. The objective is to learn to map pairs of fins from the same dolphin to be spatially close and fins from other dolphins to be spatially far.

Learning such an embedding has many useful advantages over a more traditional classification workflow in photo-id studies, where multiple images are often captured of the same individual during an encounter in the field, including:
\begin{enumerate}
\item new images from the field can be automatically grouped by individual by running a clustering algorithm, such as \textit{k}-means, on their embedding vectors;
\item new images from the field can be automatically matched to large catalogues of known individuals with a simple \textit{k}-NN classifier, or a similarity ranking can be computed and presented to the user for final verification;
\item sightings database can be efficiently checked for consistency;
\item population catalogues can be built from unlabelled images via active learning with a human-in-the-loop.
\end{enumerate}

Unlike methods that use manually engineered features \cite{Gilman2016Computer-assistedPigmentations}, our approach is completely data-driven; the model learns discriminative features that are useful at differentiating between classes (dolphin individuals), whilst being robust to intra-class variations. We posited that, given a sufficiently large and diverse training dataset, the features' discriminative power will generalise to individuals that are not present in the training data. We attempt to quantify what `sufficiently large' actually means for this kind of data.

\subsection{Triplet Loss}
\emph{Triplet loss} refers to a family of objective functions defined over triplets of images (i) an anchor, $a$, (ii) a positive example of the same class as the anchor, $p$, and (iii) a negative example of a different class, $n$. Triplet losses are typically used to optimise an embedding function $f_{\theta}(x)\! : \mathbb{R}^{data} \rightarrow \mathbb{R}^D$ that maps data points from the original high-dimensional space to a lower dimensional embedding space ($D$), such that the anchor, $a$, will be positioned closer to the positive example, $p$, than to the negative example, $n$, in $\mathbb{R}^D$, for every possible triplet $(a,p,n)$ in the training set, $X$, effectively imposing a relative distance constraint:
\begin{align*}
\begin{split}
D\left(\nnfn(x_a), \nnfn(x_p)\right) \; \leq \;  D\left(\nnfn(x_a), \nnfn(x_n)\right)
\\ \quad \forall \; (x_a,x_p,x_n) \in X
\end{split}
\end{align*}

A distance function $D(e_1, e_2)\! : \mathbb{R}^D \times \mathbb{R}^D \rightarrow \mathbb{R}_{+}$, defined on the embedding space (Euclidean or cosine distance are frequently used) can then be used as a measure of semantic similarity of the data that embedding vectors $e_1$ and $e_2$ represents. 

This motivated the triplet loss presented in \cite{Schroff2015FaceNet:Clustering}, which is the most commonly used variant:
\begin{align*}
\begin{split}
    & L(\theta; X) = \\ 
    & \sum_{(x_a, x_p, x_n) \in X}
    \big[
        D(\nnfn(x_a),  \nnfn(x_p)) - D(\nnfn(x_a),  \nnfn(x_n)) + m
    \big]^+
\end{split}
\end{align*}

Where $[x]^+\! :=\! \max(x, 0)$ and $m$ is an arbitrary positive scalar that defines the margin size. This loss function has a \emph{hard margin}: triplets with the positive example embedded significantly closer to the anchor than the negative example do not contribute to the loss. This triplet loss allows different instances of a class to lie within a local neighbourhood of the embedding space (instead of being driven to a single point) and encourages large gaps between the neighbourhoods of different classes. 

A significant drawback of the triplet loss is that it can be difficult to optimise. After some training, the model learns to map most trivial triplets in a way that they fall within the margin and stop contributing to the loss, significantly slowing down convergence. Consequently, it is important to mine for `hard' triplets that violate the hard-margin distance constraint and thereby ensure convergence (see \cite{Schroff2015FaceNet:Clustering} for more detail). 

We adopt \emph{batch-hard triplet loss} proposed by Hermans et al.~\cite{Hermans2017InRe-Identification} that samples $K$ images  from  each of $P$ individuals, stacking them into a batch of size $PK$, and uses each image as an anchor, selecting only the hardest possible triplet for each anchor to contribute to the loss.

The results presented in \cite{Hermans2017InRe-Identification} provide evidence that this hard triplet mining over a small batch, (such as $P\! =\! 21$, $K\! =\! 4$) can have the same effect on optimisation as soft triplet mining over a medium sized batch, such as that performed by FaceNet \cite{Schroff2015FaceNet:Clustering}, but is much more efficient. A key advantage is that the small batches can fit in GPU memory, allowing for both, hard triplet mining and training using stochastic gradient descent to be performed on a single machine.
 
 Following the original work, we also replaced the hard margin with a \emph{soft margin}, by changing the hinge function, $[x]^+$, to it's smooth approximation: the softplus function $\zeta(x) = \log(1+e^x)$. Soft margin has been empirically shown to work better than hard margin over a range of different values of $m$ and it's use also removes one hyperparameter. Our loss function can be expressed as:

\begin{align*}\label{eq:loss_bh}
     L(\theta; X) = \overbrace{\sum\limits_{i=1}^{P} \sum\limits_{a=1}^{K}}^{\textnormal{all anchors}}
         \zeta
         \Big(& \hspace*{-1pt} \overbrace{\max\limits_{p=1 \dots K} \hspace*{-5pt} D\left(\nnfn(x^i_a), \nnfn(x^i_p)\right)}^{\textnormal{hardest positive}} \\
                & - \hspace*{-5pt} \underbrace{\min\limits_{\substack{j=1 \dots P \\ n=1 \dots K \\ j \neq i}} \hspace*{-5pt} D\left(\nnfn(x^i_a), \nnfn(x^j_n)\right)}_{\textnormal{hardest negative}} \Big),\nonumber
 \end{align*}

\subsection{Backbone Network Architecture}

He et al. \cite{He2016DeepRecognition} present a family of convolutional neural networks that use skip layer connections to help alleviate the vanishing gradients problem; thus, allowing very deep neural networks to be trained. While deeper and wider networks with more expressive power may achieve better performance in classification, they also have a higher memory footprint, which limits the size of the mini-batch that can fit in GPU memory. Larger batch size, however, can lead to better optimisation of the triplet loss, as it allows harder triplets to be found, so there is a trade-off between model expressiveness and triplet loss optimisation. We posited that the ResNet-50 \cite{He2016DeepRecognition}, a deep residual network with 50 layers, offered a good trade-off, and used it as the backbone architecture for our experiments.

Following the work of Hermans et al. \cite{Hermans2017InRe-Identification}, we add a 1024-unit fully connected layer with the rectified linear unit (ReLU) activation function and batch normalisation \cite{Ioffe2015BatchShift}, followed by a 128 unit fully connected layer (with no activation) as the output. Projecting images onto a relatively low dimensional (128D) space has two key advantages: it helps avert the curse of dimensionality \cite{Hastie2009TheLearning} and it speeds up computation of pairwise distances on large datasets.

\subsection{Training}
All images were resized to $224 \times 224$ pixels. We used $P=21$ and $K=4$ for batch-hard triplet loss and stopped training after 2000 batches. We used the Adam optimiser \cite{Kingma2015Adam:Optimization} with the TensorFlow  default hyper-parameters ($\beta_1=0.9$, $\beta_2=0.999$ $\epsilon=10^{-8}$) to train the network. We started training with a learning rate of $3\times10^{-4}$ for 640 batches and then exponentially decayed the learning rate.  Euclidean distance was used for $D(e_1, e_2)$. All training was performed on an NVIDIA GTX-1080ti GPU.
 
Previous research has shown that initialising a network with parameters learned from another task can improve generalisation relative to a random initialisation \cite{Yosinski2014HowNetworks}. This has been key to many of the successes of DL applied to image recognition in recent years, particularly on smaller datasets such as ours.
 
We attempted to transfer learning by initialising the network weights from models pre-trained on other tasks: (i) an image classification task trained on ImageNet dataset and (ii) a person re-id task trained on Market-1501 dataset. In the second scenario, we initialised from a pre-trained model using not only the Resnet-50 weights, but also the fully connected layers that performed the embedding mapping. We considered whether a network pre-trained with triplet loss as a Euclidean embedding function may transfer better for a metric learning task on another (larger) dataset than a network pre-trained as a classifier.

\subsection{Data Augmentation}
We attempted to reduce over-fitting by performing standard data augmentation. Image hue was adjusted by a random value between 0 and 0.1 and saturation was adjusted by a factor between 0.9 and 1.1 with both values sampled from a uniform distribution. We also randomly rotated each image in the batch by an angle in degrees sampled from a truncated normal distribution with $\mu=0$ and $\sigma=5$ and maximum angle of 10. 

All training-time augmentation was performed online and no test-time augmentation was performed.

\section{EVALUATION}
Scientific studies that use photo-id depend on the accuracy of the method. Computer vision algorithms, such as the one proposed here, can be used to assist researchers in their cataloguing task, but the final verification must be performed by a human. So the problem can be formulated as image retrieval: given a query image of an unknown individual, an algorithm matches it to existing images in the catalogue of known individuals and returns the top $k$ identities, ranked by similarity; a human can then assess these and pick the correct match. A good algorithm will have a high chance of returning the correct identity as the most likely (top-1) result; however, even having the correct identity in the top 5 results will still significantly reduce the number of comparisons a human would need to perform.

To evaluate the ranking performance of our model, we split the evaluation/test set (described in \ref{sec:cv}) into a query image and a gallery set. We ranked gallery images by similarity to the query image and computed top-$k$ accuracy (for $k\in\{1,5\}$) and mean average precision (mAP) across all the query/gallery set combinations.

Many of the dolphins in our dataset have multiple images of them taken on the same day. These images have very strong correlations (due to the same camera being used under similar lighting and weather conditions) which can lead to artificially inflated performance figures. To mitigate this, we excluded from the gallery set, all images of the query individual that were taken on the same day as the query image. Additionally, we did not perform multiple queries from the same day/individual combination, instead we randomly selected a single query image from all images of the same individual taken on the same day.

\subsection{Data}
Results are based on 3544 cropped and labelled images from 185 individual dolphins. Images were collected over multiple years (between 2002 and 2014) under a wide variety of weather/lighting conditions. No alignment of any kind was performed at any stage.

\label{sec:cv}
\subsection{Cross-validation}
We ran 5-fold cross-validation after the dataset was sorted by dolphin identity and ordered using the number of images-per-dolphin. Each fold was created by systematically selecting every $5^{th}$ dolphin (with a different starting position).  As a result, each fold consisted of 37 individuals with approximately the same distribution of images-per-dolphin. One fold was used as the evaluation set, with the remaining four as the training set.

We computed the mean and standard error for every metric across each fold being used as the evaluation set.

\subsection{Distractors}
To show how performance metrics scaled to large catalogues containing hundreds or even thousands of unique individuals, we progressively added distractor images to the gallery. The gallery set for each query contained, on average, 700 images and we added up to 1200 distractors, for a maxmimum total of 1900 images.

\subsection{Training set size}
Knowing how the size of the training set affects performance is useful because it gives an idea of how much adding more labelled dolphins will improve performance. It may also shed light on the amount of labelled data required to train a useful embedding for new species. We picked a single fold of 37 individuals to be a fixed evaluation set for this investigation and repeatedly trained our model on random subsets of dolphins from the remaining four folds.

We performed two experiments using the ResNet-50 model pre-trained on ImageNet with data augmentation: (i) we systematically reduced the number of individuals in the training set and (ii) we used all 148 individuals, but systematically reduced the maximum number of images per individual.

\section*{Results \& Discussion}
We visualised and examined the clustering of embedding vectors using t-SNE \cite{Maaten2008VisualizingT-SNE}. Tight clusters, corresponding closely to dolphin identity, could be observed in the training set. However, the evaluation set (pictured in Fig. \ref{fig:t-sne}) exhibited (relatively) less correspondence between clusters and identities, suggesting that the model may have been overfitting. The model seem to be able to extract discriminative features that generalise to new identities, but a better augmentation strategy (such as incorporating perspective or affine transformation) may be required to reduce overfitting even further.

\begin{table}[!h]
\caption{5-fold cross-validation results. Mean $\pm$ SE}
\label{tab:results}
\centering
\begin{tabular}{|l|l|l|l|}
\hline
           & mAP        & Top 1     & Top 5     \\ \hline
Market-1501 init & $64.9\pm1$  & $80.8 \pm 4$ & $90.1 \pm 1$ \\ \hline
ImageNet init   & $76 \pm 1$    & $85.2 \pm 2$ & $92.9 \pm 2$ \\ \hline
ImageNet + Augment    & $80.8 \pm 1$ & $90.6 \pm 2$  & $93.6 \pm 1$  \\ \hline
\end{tabular}
\end{table}

Table \ref{tab:results} shows results for ResNet-50 pre-trained on ImageNet, and ResNet-50 pre-trained on a person re-id task on Market-1501 dataset with triplet loss. Both performed better than completely random initialisation. ImageNet initialisation achieved significantly better performance than Market-1501 (76 vs 65 mAP), even though it was trained as a softmax classifier with cross-entropy loss and fully connected embedding mapping layers were initialised randomly, in contrast to all layers initialised using a model pre-trained using triplet-loss on Market-1501. Better performance of ImageNet initialisation over random was expected, as it contains many visually similar images to our data, such as whales and sharks---so ImageNet features should be useful for dolphin recognition tasks. However, it was surprising that Market-1501 trained model did not perform as well, considering that it was itself initialised with ImageNet-trained weights, prior to being trained on Market-1501 with triplet-loss. We hypothesise that catastrophic forgetting plays a role in this performance degradation, but investigating it further was outside the scope of this work. 

\begin{figure*}[t!]
\centering
\vspace*{-.3cm}
\includegraphics[width=0.66\textwidth]{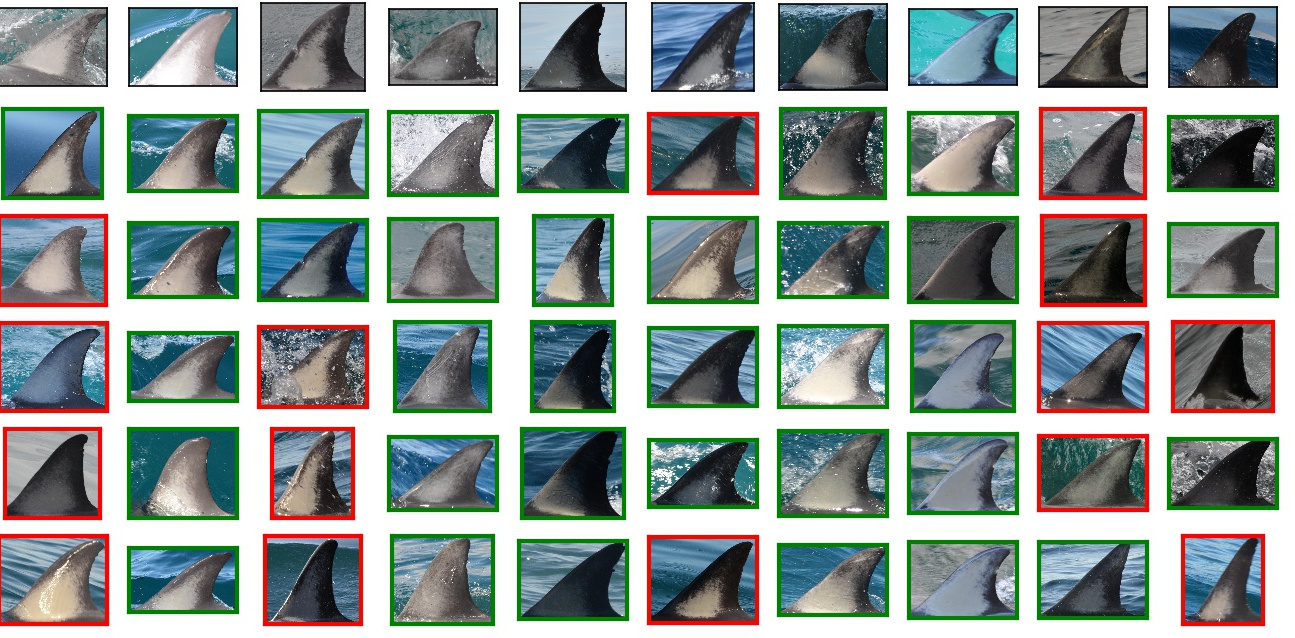}
\vspace*{-.2cm}
\caption{A random sample of similarity rankings. Each column contains the query image at the top and returned results below it. Images are ranked by decreasing similarity top to bottom. A green border indicates correct match and red indicates a mismatch.}
\vspace*{-.2cm}
\label{fig:fin-rankings}
\end{figure*}

 A model, initialised from ImageNet, and trained with data augmentation gives us the best results for every metric: $80.8\pm1$ mAP, $90.6\pm2$ percent top-1 accuracy and $93.6\pm1$ percent top-5 accuracy.  Fig \ref{fig:fin-rankings} shows a random selection of query images and their returned results.
 
\begin{figure}
\centering
\includegraphics[width=\columnwidth]{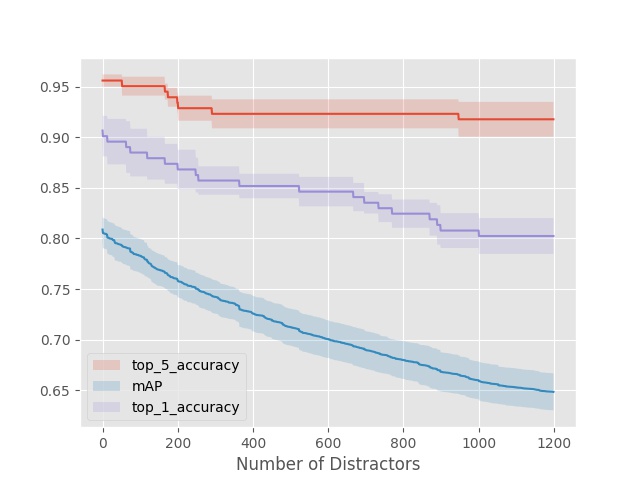}
\caption{Performance vs number of distractor images in the evaluation set. The shaded area represents the standard error calculated across different folds.
Note that the validation set before adding distractors had an average size of around 700.}
\label{fig:distractors}
\end{figure}

 In the presence of distractors, we observe an approximately linear degradation in performance (see Fig. \ref{fig:distractors}). The performance of this model drops to $64.8\pm2$ mAP, $80\pm2$ top-1 and $91\pm2$ top-5 in the presence of 1200 distractors. A model with such accuracy would still be very useful in practice: being able to pick the correct match out of a list of 5 individuals (instead of a list of 1237) in 9 out of 10 times would be a great improvement for the photo-id workflow.

\begin{figure}
\centering
\includegraphics[width=\columnwidth]{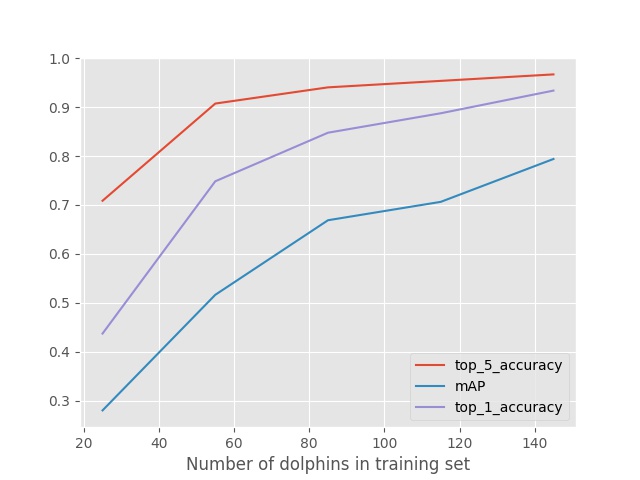}
\caption{Performance vs number of individual dolphins in the training set.}
\label{fig:reduced_individuals_training_set}
\end{figure}

\begin{figure}
\centering
\vspace*{-0.5cm}
\includegraphics[width=\columnwidth]{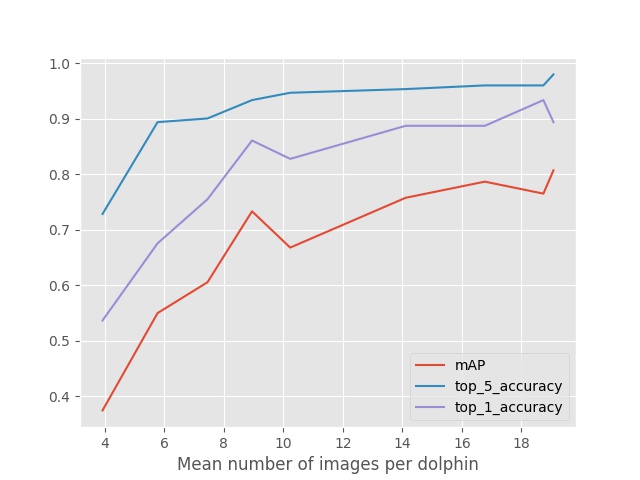}
\caption{Performance vs mean number of images per individual in training set.}
\label{fig:reduced_images_training_set}
\end{figure}

Fig. \ref{fig:reduced_individuals_training_set} shows the result of varying the number of individuals in the training set. It appears that adding more individuals to the training set improves generalisation albeit with diminishing returns. We can still achieve reasonable top-5 accuracy of ~90 percent with only 60 individuals.

Fig. \ref{fig:reduced_images_training_set} shows the result of capping the maximum number of images per individual in the training set. The results appear to have high variance, due to our method of selecting a random subset of images that are used in the training set for each dolphin, instead of using a random subset of the training set with larger cap. However, the general trend is visible; we need at least 8--10 images per individual to get reasonable top-5 performance ($>$80\% accuracy).

We suspect diminishing improvements are exacerbated by the presence of multiple images of the same dolphin captured during the same encounter (same day) in the training set and believe that including more images captured under different conditions would improve the results even further.

We also hypothesise that there is a cross-dependence between number of individuals and number of images per individual, however, we did not have the resources to verify this claim, but will attempt this in the future.

\section{CONCLUSIONS}
Our approach of using a deep convolutional network to learn an embedding provided classification success rates that were substantially higher than attempts that utilised hand-engineered features of common dolphin images \cite{Pawley2018ExaminingDelphinids, Gilman2016Computer-assistedPigmentations}. Moreover, the previous attempts also required considerable pre-processing of the images, including manual segmentation of the fin from the background and alignment of each fin to a canonical exemplar using the ICP algorithm \cite{Gilman2013DolphinPhoto-identification}. The use of an embedding naturally generalises to images of new dolphins, since it makes no \textit{a priori} assumption that all classses are known. Future work aims to generalise the model to include other dolphin species (e.g. bottlenose dolphins) and produce an application that will ease the ecologists' catalogue workflow.

\bibliography{dolphins}

\begin{thebibliography}{10}
\providecommand{\url}[1]{#1}
\csname url@samestyle\endcsname
\providecommand{\newblock}{\relax}
\providecommand{\bibinfo}[2]{#2}
\providecommand{\BIBentrySTDinterwordspacing}{\spaceskip=0pt\relax}
\providecommand{\BIBentryALTinterwordstretchfactor}{4}
\providecommand{\BIBentryALTinterwordspacing}{\spaceskip=\fontdimen2\font plus
\BIBentryALTinterwordstretchfactor\fontdimen3\font minus
  \fontdimen4\font\relax}
\providecommand{\BIBforeignlanguage}[2]{{%
\expandafter\ifx\csname l@#1\endcsname\relax
\typeout{** WARNING: IEEEtran.bst: No hyphenation pattern has been}%
\typeout{** loaded for the language `#1'. Using the pattern for}%
\typeout{** the default language instead.}%
\else
\language=\csname l@#1\endcsname
\fi
#2}}
\providecommand{\BIBdecl}{\relax}
\BIBdecl

\bibitem{Wursig1977photographic}
B.~W{\"{u}}rsig and M.~W{\"{u}}rsig, ``{The Photographic Determination of Group
  Size, Composition, and Stability of Coastal Porpoises (Tursiops
  truncatus)},'' \emph{Science}, vol. 198, no. 4318, pp. 755--756, 1977.

\bibitem{Pawley2018ExaminingDelphinids}
M.~D.~M. Pawley, K.~E. Hupman, K.~A. Stockin, and A.~Gilman, ``{Examining the
  viability of dorsal fin pigmentation for individual identification of
  poorly-marked delphinids},'' \emph{Scientific Reports}, vol.~8, no.~1, p.
  12593, 12 2018.

\bibitem{Hupman2018ChallengesDelphinids}
K.~Hupman, K.~A. Stockin, K.~Pollock, M.~D.~M. Pawley, S.~L. Dwyer, C.~Lea, and
  G.~Tezanos-Pinto, ``{Challenges of implementing Mark-recapture studies on
  poorly marked gregarious delphinids},'' \emph{PLOS ONE}, vol.~13, no.~7, p.
  e0198167, 7 2018.

\bibitem{LeCun2015DeepLearning}
Y.~A. LeCun, Y.~Bengio, and G.~E. Hinton, ``{Deep learning},'' \emph{Nature},
  2015.

\bibitem{Krizhevsky2012ImageNetNetworks}
A.~Krizhevsky, I.~Sutskever, and G.~E. Hinton, ``{ImageNet Classification with
  Deep Convolutional Neural Networks},'' \emph{Advances In Neural Information
  Processing Systems}, 2012.

\bibitem{Ren2017FasterNetworks}
S.~Ren, K.~He, R.~Girshick, and J.~Sun, ``{Faster R-CNN: Towards Real-Time
  Object Detection with Region Proposal Networks},'' \emph{IEEE Transactions on
  Pattern Analysis and Machine Intelligence}, 2017.

\bibitem{Kuhl2013AnimalAppearance}
H.~S. K{\"{u}}hl and T.~Burghardt, ``{Animal biometrics: Quantifying and
  detecting phenotypic appearance},'' 2013.

\bibitem{Zheng2016PersonFuture}
\BIBentryALTinterwordspacing
L.~Zheng, Y.~Yang, and A.~G. Hauptmann, ``{Person Re-identification: Past,
  Present and Future},'' \emph{arXiv preprint arXiv:1610.02984}, 10 2016.
  [Online]. Available: \url{http://arxiv.org/abs/1610.02984}
\BIBentrySTDinterwordspacing

\bibitem{Scheirer2013TowardRecognition}
W.~J. Scheirer, A.~de~Rezende~Rocha, A.~Sapkota, and T.~E. Boult, ``{Toward
  Open Set Recognition},'' \emph{IEEE Transactions on Pattern Analysis and
  Machine Intelligence}, vol.~35, no.~7, pp. 1757--1772, 7 2013.

\bibitem{Taigman2014DeepFace:Verification}
Y.~Taigman, M.~Yang, M.~Ranzato, and L.~Wolf, ``{DeepFace: Closing the Gap to
  Human-Level Performance in Face Verification},'' in \emph{2014 IEEE
  Conference on Computer Vision and Pattern Recognition}.\hskip 1em plus 0.5em
  minus 0.4em\relax IEEE, 6 2014, pp. 1701--1708.

\bibitem{Sun2015DeeplyRobust}
Y.~Sun, X.~Wang, and X.~Tang, ``{Deeply learned face representations are
  sparse, selective, and robust},'' in \emph{Proceedings of the IEEE Computer
  Society Conference on Computer Vision and Pattern Recognition}, vol.
  07-12-June, 12 2015, pp. 2892--2900.

\bibitem{Schroff2015FaceNet:Clustering}
F.~Schroff, D.~Kalenichenko, and J.~Philbin, ``{FaceNet: A unified embedding
  for face recognition and clustering},'' in \emph{Proceedings of the IEEE
  Computer Society Conference on Computer Vision and Pattern Recognition}, vol.
  07-12-June, 3 2015, pp. 815--823.

\bibitem{Tadmor2016LearningMethod}
O.~Tadmor, Y.~Wexler, T.~Rosenwein, S.~Shalev-Shwartz, and A.~Shashua,
  ``{Learning a Metric Embedding for Face Recognition using the Multibatch
  Method},'' \emph{Advances in Neural Information Processing Systems}, 2016.

\bibitem{Hermans2017InRe-Identification}
\BIBentryALTinterwordspacing
A.~Hermans, L.~Beyer, and B.~Leibe, ``{In Defense of the Triplet Loss for
  Person Re-Identification},'' \emph{arXiv preprint arXiv:1703.07737}, 3 2017.
  [Online]. Available: \url{http://arxiv.org/abs/1703.07737}
\BIBentrySTDinterwordspacing

\bibitem{Parkhi2015DeepRecognition}
O.~M. Parkhi, A.~Vedaldi, and A.~Zisserman, ``{Deep Face Recognition},'' in
  \emph{Procedings of the British Machine Vision Conference 2015}.\hskip 1em
  plus 0.5em minus 0.4em\relax British Machine Vision Association, 2015, pp.
  1--41.

\bibitem{PhotoApplication}
\BIBentryALTinterwordspacing
``{Photo ID Ninja - Dolphin Recognition Web Application}.'' [Online].
  Available: \url{http://photoid.ninja/}
\BIBentrySTDinterwordspacing

\bibitem{Szegedy2016RethinkingVision}
C.~Szegedy, V.~Vanhoucke, S.~Ioffe, J.~Shlens, and Z.~Wojna, ``{Rethinking the
  Inception Architecture for Computer Vision},'' in \emph{2016 IEEE Conference
  on Computer Vision and Pattern Recognition (CVPR)}.\hskip 1em plus 0.5em
  minus 0.4em\relax IEEE, 6 2016, pp. 2818--2826.

\bibitem{Gilman2016Computer-assistedPigmentations}
A.~Gilman, K.~Hupman, K.~A. Stockin, and M.~D.~M. Pawley, ``{Computer-assisted
  recognition of dolphin individuals using dorsal fin pigmentations},'' in
  \emph{2016 International Conference on Image and Vision Computing New Zealand
  (IVCNZ)}.\hskip 1em plus 0.5em minus 0.4em\relax IEEE, 11 2016, pp. 1--6.

\bibitem{He2016DeepRecognition}
K.~He, X.~Zhang, S.~Ren, and J.~Sun, ``{Deep Residual Learning for Image
  Recognition},'' in \emph{2016 IEEE Conference on Computer Vision and Pattern
  Recognition (CVPR)}, 2016.

\bibitem{Ioffe2015BatchShift}
S.~Ioffe and C.~Szegedy, ``{Batch Normalization: Accelerating Deep Network
  Training by Reducing Internal Covariate Shift},'' in \emph{Proceedings of the
  32nd International Conference on International Conference on Machine Learning
  - Volume 37}, 2015, pp. 448--456.

\bibitem{Hastie2009TheLearning}
T.~Hastie, R.~Tibshirani, and J.~Friedman, \emph{{The Elements of Statistical
  Learning}}, 2009.

\bibitem{Kingma2015Adam:Optimization}
D.~P. Kingma and J.~L. Ba, ``{Adam: a Method for Stochastic Optimization},''
  \emph{International Conference on Learning Representations}, 2015.

\bibitem{Yosinski2014HowNetworks}
J.~Yosinski, J.~Clune, Y.~Bengio, and H.~Lipson, ``{How Transferable Are
  Features in Deep Neural Networks?}'' in \emph{Proceedings of the 27th
  International Conference on Neural Information Processing Systems - Volume
  2}, 2014, pp. 3320--3328.

\bibitem{Maaten2008VisualizingT-SNE}
L.~v.~d. Maaten and G.~Hinton, ``{Visualizing Data using t-SNE},''
  \emph{Journal of Machine Learning Research}, vol.~9, no. Nov, pp. 2579--2605,
  2008.

\bibitem{Gilman2013DolphinPhoto-identification}
A.~Gilman, T.~Dong, K.~Hupman, K.~Stockin, and M.~Pawley, ``{Dolphin fin pose
  correction using ICP in application to photo-identification},'' in
  \emph{International Conference Image and Vision Computing New Zealand}, 2013,
  pp. 388--393.

\end{thebibliography}
\bibliographystyle{IEEEtran}

\end{document}